\documentclass{article}

\usepackage[preprint]{neurips_2022}

\usepackage[utf8]{inputenc} 
\usepackage[T1]{fontenc}    

\usepackage{url}            
\usepackage{booktabs}       
\usepackage{amsfonts}       
\usepackage{nicefrac}       
\usepackage{microtype}      
\usepackage{xcolor}         
\usepackage{amsmath}
\usepackage{amssymb}
\usepackage{algorithm}
\usepackage{algorithmic}
\usepackage{subfigure}
\usepackage{graphicx}
\usepackage{multirow}
\usepackage{color}
\usepackage{makecell}
\usepackage{hyperref}       

\title{TRT-ViT: TensorRT-oriented Vision Transformer}

\author{%
    Xin Xia\footnotemark[1],\qquad Jiashi Li\footnotemark[1] ,\qquad Jie Wu\footnotemark[2] ,\qquad Xing Wang\footnotemark[2], \\
    \textbf{Xuefeng Xiao,\qquad Min Zheng,\qquad Rui Wang}\\
 ByteDance Inc \\
 	{\tt\small \{xiaxin.97, lijiashi, wujie.10, wangxing.613\}@bytedance.com}\\
 	 	{\tt\small \{xiaoxuefeng.ailab, zhengmin.666, ruiwang.rw\}@bytedance.com}
}

\begin{document}

\maketitle

\renewcommand{\thefootnote}{\fnsymbol{footnote}} 
\footnotetext[1]{Equal contribution.}
\footnotetext[2]{Corresponding authors are Xing Wang and Jie Wu.}

	\begin{abstract}
		We revisit the existing excellent Transformers from the perspective of practical application. Most of them are not even as efficient as the basic ResNets series and deviate from the realistic deployment scenario. It may be due to the current criterion to measure computation efficiency, such as FLOPs or parameters is \textit{one-sided, sub-optimal, and hardware-insensitive}. Thus, this paper directly treats the TensorRT latency on the specific hardware as an efficiency metric, which provides more comprehensive feedback involving computational capacity, memory cost, and bandwidth. Based on a series of controlled experiments, this work derives four practical guidelines for TensorRT-oriented and deployment-friendly network design, e.g., early CNN and late Transformer at stage-level, early Transformer and late CNN at block-level. Accordingly, a family of TensortRT-oriented Transformers is presented, abbreviated as TRT-ViT. Extensive experiments demonstrate that TRT-ViT significantly outperforms existing ConvNets and vision Transformers with respect to the latency/accuracy trade-off across diverse visual tasks, e.g., image classification, object detection and semantic segmentation. For example, at 82.7\% ImageNet-1k top-1 accuracy, TRT-ViT is \textbf{2.7$\times$} faster than CSWin and \textbf{2.0$\times$}  faster than Twins. On the MS-COCO object detection task, TRT-ViT achieves comparable performance with Twins, while the inference speed is increased by \textbf{2.8$\times$}.
		
	\end{abstract}

	\section{Introduction}
	
	
	
	Recently, Vision Transformer has witnessed prevailing success and achieved noticeable performance gain over CNN in various computer vision tasks, such as image classification, semantic segmentation, and object detection, etc. 
	However, CNN still dominates the visual architecture from the perspective of real-scene deployment.
	When we dig deep into substantial performance improvements behind existing Transformers, we observe that these gains come at the cost of extensive resource overhead.
	Moreover, it has become a demanding obstacle when the Transformer is deployed on resource-constrained devices. 
	To alleviate unwieldy computational cost of Transformer, a series of works pay much attention to developing efficient Vision Transformer, such as Swin Transformer \cite{Swin}, PVT \cite{PVT_v1}, Twins \cite{Twins}, CoAtNet\cite{Coatnet} and MobileViT\cite{MobileViT}.
	
	\begin{figure*}[htbp]
		\centering
		\subfigure[] {
			\label{fig:a}
			\includegraphics[width=0.3\columnwidth]{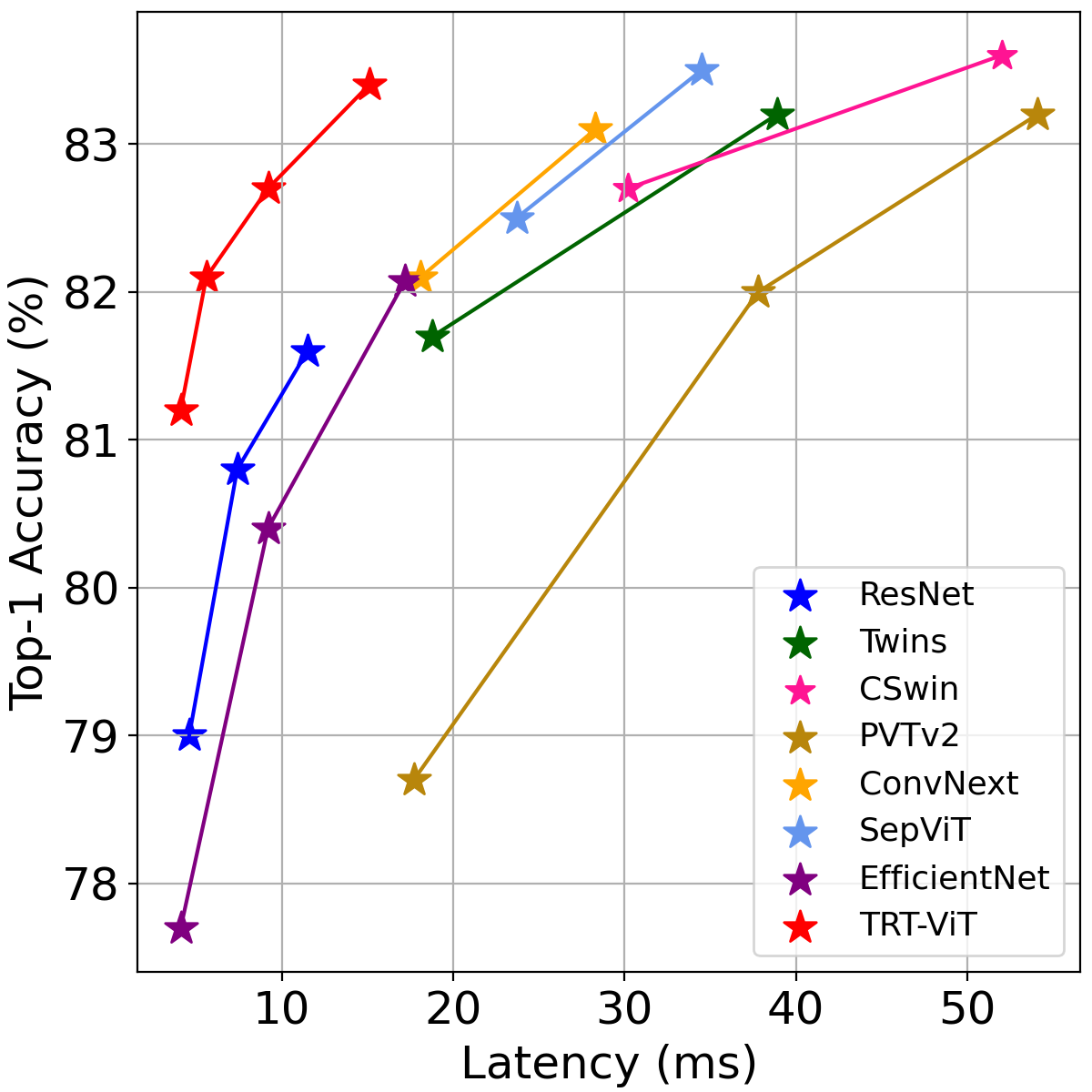}
		}
		\hspace{-0.1in} 
		\subfigure[] {
			\label{fig:b}
			\includegraphics[width=0.3\columnwidth]{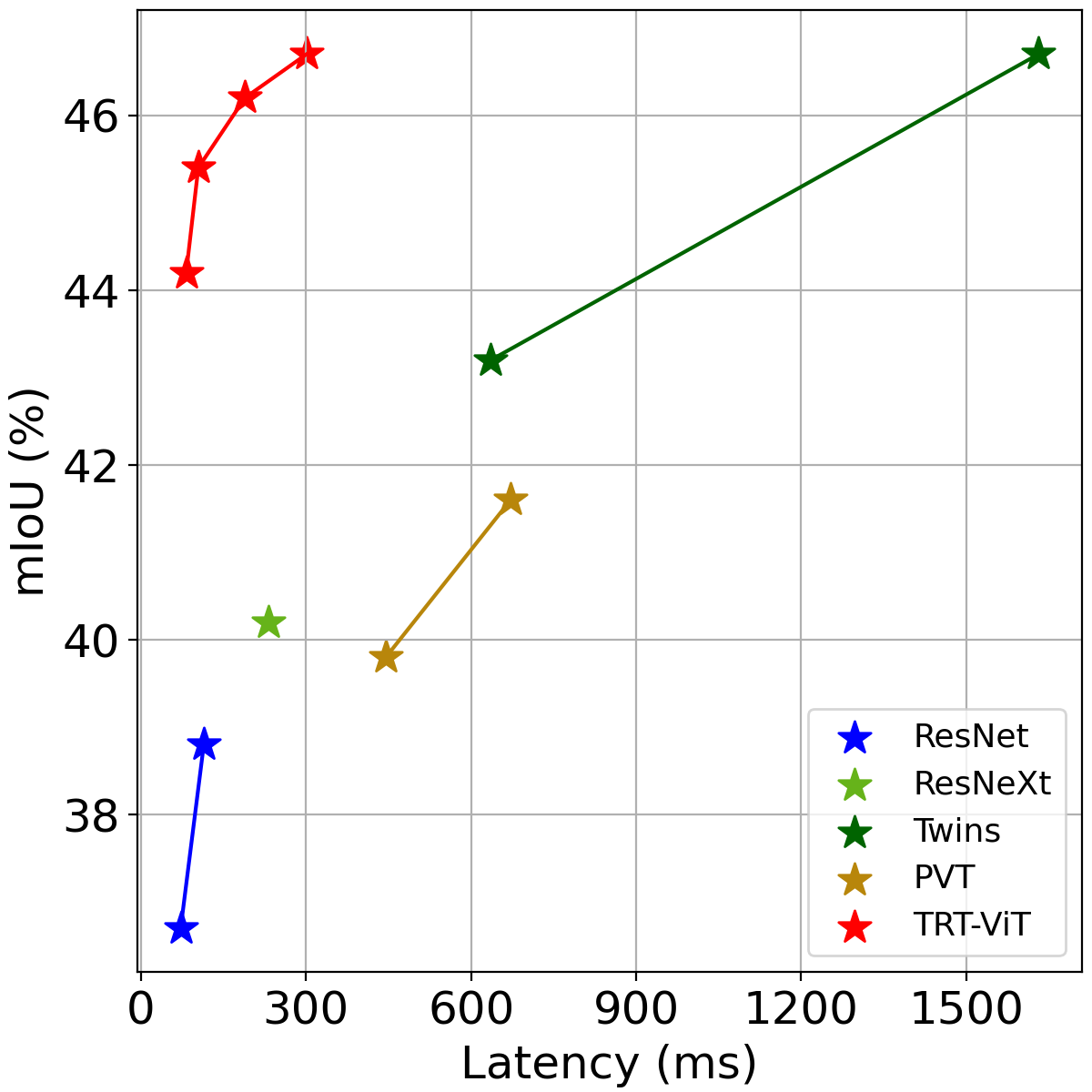}
		}
		\hspace{-0.1in} 
		\subfigure[] {
			\label{fig:b}
			\includegraphics[width=0.3\columnwidth]{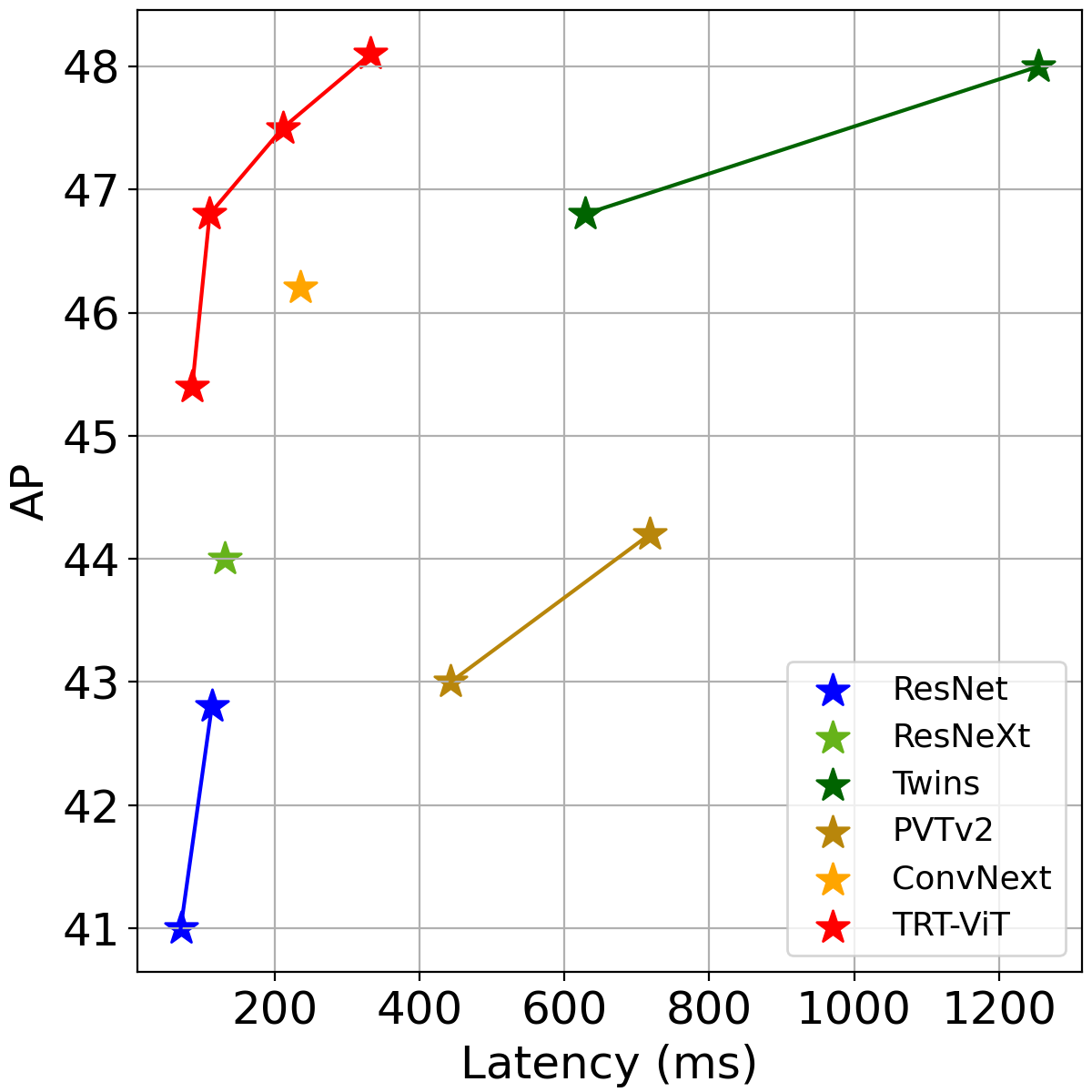}
		}
		
		\caption{Comparison among TRT-ViT and efficient Networks, in terms of accuracy-latency trade-off. (a): ImageNet classification; (b): ADE20K Semantic Segmentation; (c): COCO Object Detection.  TRT-ViT consistently outperforms the SOTA efficient Networks.  Latency is measured on the TensorRT framework with a T4 GPU (batch size=8)}
		\label{result}
 		\vspace{-0.6cm}
	\end{figure*}
	Although the above works showed the efficiency of ViTs from different points of view, the critical design is mainly guided by the indirect criterion of computation efficiencies, such as FLOPs or parameters.
	This circuitous criterion is one-sided, sub-optimal, and deviating from real-scene deployment.
	In fact, the model has to cope with environmental uncertainty in the deployment procedure, which involve hardware-aware characteristics such as memory access cost and I/O throughput.
	Specifically, as stated in many papers (e.g.,\cite{ShuffleNet_v2, GPUNet}), the indirect metric of computation complexity is widely used in neural network architecture design that FLOPs and parameters are approximations of, but usually can not truly reflect the direct metric that we really care about, such as speed or latency.
	In this paper, to address such discrepancy, we provide a view that treats the TensorRT latency on the specific hardware as direct efficiency feedback.
	TensorRT has become a general and deployment-friendly solution in practical scenarios, and it helps to provide convincing hardware-oriented guidance.
	With this direct and accurate guidance, we redraw the accuracy and efficiency trade-off diagram of several existing competitive models in Figure~\ref{result}. As depicted in Figure~\ref{result}, Transformer has the advantage of good performance, while CNN successes in high efficiency.
	Although ResNet is not as impressive as the competitive Transformers in performance, it is still the best architecture under the accuracy-latency trade-off. For example, ResNet employs 11.7 milliseconds (batch size = 8)  to achieve 81.7\% accuracy. Although twins-pcpvt-s achieves an impressive accuracy, 83.4\% in imagenet-1k classification, it takes up to 39.8 milliseconds to achieve this goal.
	These observations motivate us to raise a question: 
	\textit{how to design a model that can perform as well as Transformer and predict as fast as ResNet?}
	
	To answer this question, we systematically explore the hybrid design of CNNs and Transformers. We follow the stage-to-block hierarchical roadmap to investigate the TensorRT-oriented architecture.
	Through extensive experiments, we summarize four practical guidelines to design efficient networks on TensorRT: 
	1) stage-level: Transformer block at the late stages maximizes the trade-off between efficiency and performance; 
	2) stage-level: shallow-then-deep stage pattern improves the performance;
	3) block-level: a block mixed of Transformer and BottleNeck is more efficient than Transformer.; 
	4) block-level: global-then-local block pattern helps make up the performance issue.

	Based on the above guidelines,  we design a family of TensortRT-oriented Transformers (abbreviated as TRT-ViT), consisted of  hybrid networks with ConvNets and Transformers.
	CNN and Transformer are position-complementary in stage and block-level at TRT-ViT.
	Furthermore, we propose various TRT-ViT blocks to combine CNN and Transformer in a serial scheme that decouples heterogeneous concepts and improves the efficiency of information flow.
	The proposed TRT-ViT outperforms existing ConvNets and vision Transformers with respect to the latency-accuracy trade-off on image classification. For example, at 82.7\% ImageNet-1k top-1 accuracy, TRT-ViT is \textbf{2.7$\times$} faster than CSWin and \textbf{2.0$\times$}  faster than Twins.
	
	More importantly, TRT-ViT shows a more significant latency/accuracy trade-off gain on downstream tasks.
	As depicted in Figure ~\ref{result}, TRT-ViT achieves comparable performance with Twins, while the inference speed is increased by \textbf{2.8$\times$}  on the MS-COCO object detection and \textbf{4.4$\times$} on ADE20K semantic segmentation task.
	TRT-ViT outperforms ResNet by \textbf{3.2 AP} (from 40.4 to 43.6) on COCO detection and \textbf{6.6\%} (from 38.86\% to 45.46\%) on ADE20K segmentation under similar TensorRT latency. 
	
	We believe that our work points out a viable path for scientific research towards industrial deployment in terms of architecture design, which significantly reduces the gap between laboratory research and real-scene practice.
	We hope that our work can provide inspiring insight and encourage more researchers to be involved in real-scene architecture design.

	\begin{figure}[h]
		\centering
		\includegraphics[scale=0.3]{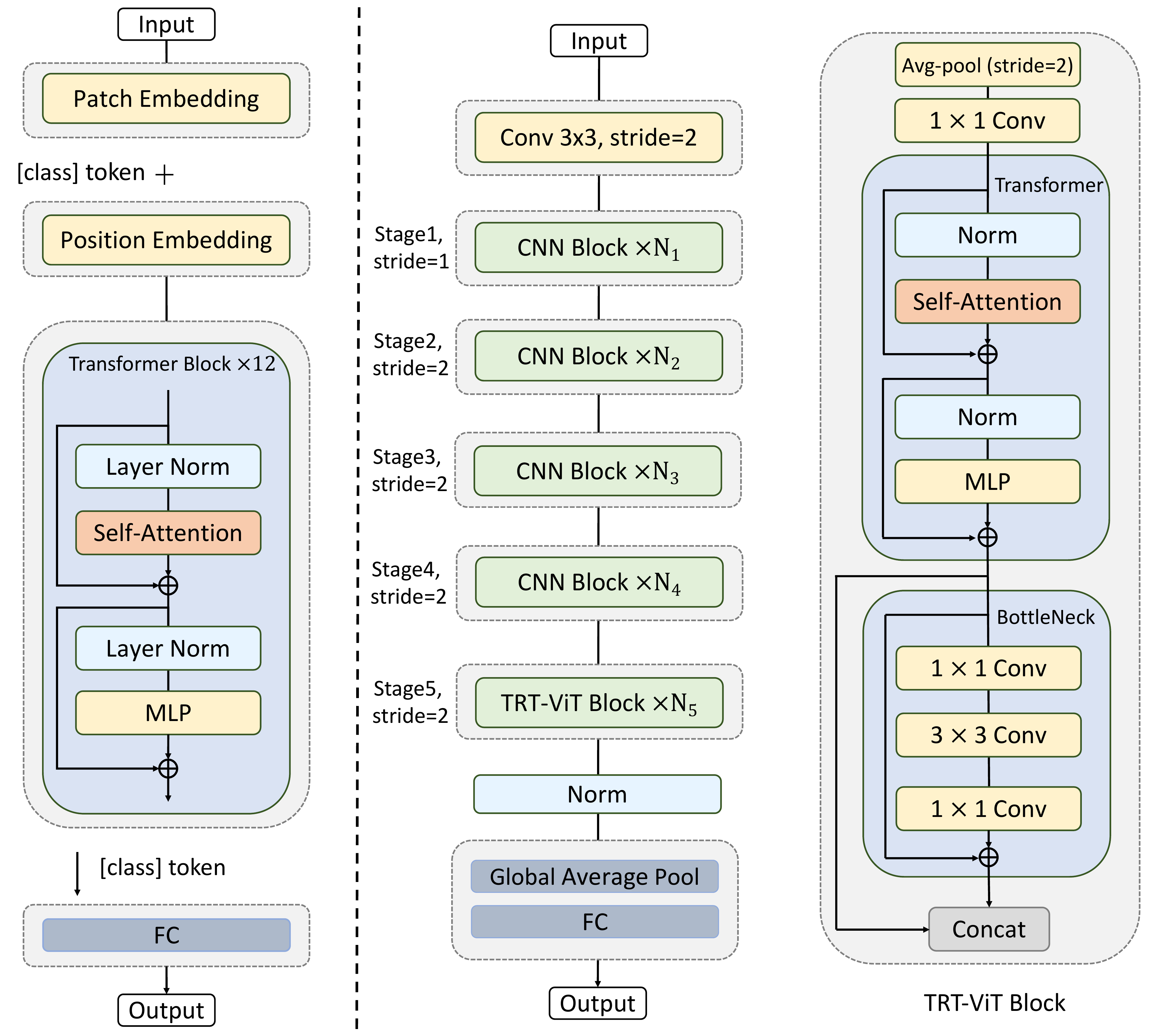}
		\caption{The overall framework TRT-ViT. \textbf{Left:} ViT model. \textbf{Right:} Our proposed TRT-ViT model.}
		\label{figure:structure}
		 		\vspace{-0.2cm}
	\end{figure}
	
	\section{Related Work}
	\textbf{Convolutional networks. }
	Convolutional neural networks (CNNs) have dominated the visual architecture in various computer vision tasks in the past decade, including image classification, object detection, semantic segmentation.
	ResNet\cite{ResNet} employs residual connections to relieve the network degradation, ensuring the network builds deeper to capture high-level abstraction. 
	DenseNet\cite{DenseNet} alternatively enhances feature reuse and connects the feature maps via dense connections. 
	MobileNets\cite{Mobilenets, MobileNet_v2}  introduces depthwise convolutions and pointwise convolutions to enhance model efficiency.
	ShuffleNet\cite{Shufflenet, ShuffleNet_v2}  adopts group convolution and channel shuffle to reduce the computational cost further. ConvNext\cite{ConvNext} reviews the design of the vision Transformers and reveals that the CNN-only architecture also can achieve comparable results with Transformers.	
	
	\textbf{Vision Transformers. }
	Transformers are first proposed in the field of natural language processing. Vision Transformer(ViT)\cite{ViT} shows that Transformer can also achieve impressive performance on multiple vision tasks. ViT splits the image into patches and treats the patches as words to perform self-attention. DeiT\cite{Deit} proposes the hierarchical Transformers to handle high-resolution images. T2T-ViT\cite{T2T} gradually converts the image into a token by recursively aggregating adjacent tokens. Swin Transformer\cite{Swin} computes attention within the shifted window.  PiT\cite{PiT} jointly uses pooling layer and depthwise convolution layer to achieve channel multiplication and spatial reduction. Nowadays, researchers pay more attention to efficiency, including efficient self-attention, training strategies, pyramidal designs, etc.
	
	\textbf{Hybrid Models. }
	Recent works show that combining convolution and Transformer as a hybrid architecture helps absorb the advantages of both architectures.
	BoTNet\cite{BoTNet} replaces the last three blocks of ResNet with a self-attention block, and CvT\cite{CvT} introduces depthwise and pointwise convolution in front of self-attention. The CMT\cite{Cmt} block consists of depthwise convolution based local perception unit and a light-weight transformer module. LeViT\cite{LeViT} substitutes the patchy stem with a convolutional stem to reduce computation cost.
	 In MobileViT\cite{MobileViT}, MobileNetV2 block and Transformer block are combined in series and a MobileVit block is developed to learn global representations. But MobileVit is still relatively heavy compared to Mobile CNNs.
	 Mobile-Former\cite{Mobile-Former} builds bidirectional fusion between Transformer and Mobilenets to capture local and holistic concepts. 
	In this paper, we design a family of TensorRT-oriented hybrid models that adapt more to the actual development scenarios.

	\section{Approach}
	In this section, we first present a series of empirical findings and propose four practical guidelines for designing efficient networks on TensorRT.  Then we develop a new architecture with high efficiency and performance, denoted as TRT-ViT. 
	
	\subsection{Practical Guidelines for Efficient Network Design on TensorRT}
	Our study is performed on the widely adopted high-performance inference SDK, TensorRT. To initiate our study, we analyze the runtime performance of two SOTA networks, ResNet \cite{ResNet} and ViT \cite{ViT}. They are both highly accurate on the ImageNet1K classification task and prevalent in the industry. Although we only analyze these two networks, we note that they are representative of the current trend. At their core are BottleNeck block and Transformer block, which are also crucial components for other state-of-the-art networks, such as ConvNext \cite{ConvNext}, and Swin \cite{Swin}.  
	
	To better illustrate our empirical results, follow RepLKNet\cite{RepLKNet}, we use the computational density (measured in Tera FLoating-point Operations Per Second, TeraFLOPS) of an operation or block to quantify its efficiency on a hardware, described as follows
	\begin{equation}
	\text{TeraFLOPS} = \frac{\text{FLOPs}}{\text{Latency}}
	\label{eq:TeraFLOPS}
	\end{equation}
	
	As seen from Figure~\ref{result}(a) and Table\ref{tab:Complexities}, BottleNeck block is more efficient and has a higher TeraFLOPS, while Transformer block is more powerful and has better accuracy on ImageNet. In this paper, we aim to design a network that achieves ResNet-like efficiency and Transformer-like performance. It is straightforward to design a network that is mixed of BottleNeck and Transformer block, e.g., MobileViT\cite{MobileViT}. However, these papers either focus on optimizing FLOPs, which is an indirect criterion to measure the efficiency, or put much effort into mobile devices. 
	
	Trivially applying Transformer to CNNs usually leads to inferior performance and speed. Hence, we summarize four practical guidelines for effectively designing efficient networks on TensorRT.

	\begin{table}[h]
		\centering
		\caption{Analyze TeraFLOPS on TensorRT. Latency is measured on the TensorRT framework with a T4 GPU (batch size=16). }
		\resizebox{0.9\textwidth}{!}{
			\begin{tabular}{c|c|c|c|c|c|c}
				\toprule
				Block Type & Feature map size& Params(K) & FLOPs(M) & Latency(ms) &TeraParams($\uparrow$) & TeraFLOPS($\uparrow$)  \\ \midrule
				\multirow{4}{*}{Transformer}  
				
				& 256$\times$56$\times$56  &658 & 7098  & 86.9  &7.5 & 81   \\ 
				& 512$\times$28$\times$28  &2627 & 2688  & 14.6 & 179& 184   \\ 
				& 1024$\times$14$\times$14 &10498 & 2135  & 4.92 &2142 & 435   \\ 
				& 2048$\times$7$\times$7 &41960 & 2066  & 3.51 &\textbf{11988} & \textbf{599}  \\ \midrule
				
				\multirow{4}{*}{BottleNeck}
				&256$\times$56$\times$56 &70 & 218         & 0.58   & 120     & 378     \\ 
				& 512$\times$28$\times$28 & 279 & 218      & 0.38   &733     & 573             \\ 
				& 1024$\times$14$\times$14  &1114 & 218    & 0.35   & 3173   & 620               \\ 
				& 2048$\times$7$\times$7    &4456 & 218   & 0.33   &\textbf{13723}  &\textbf{670}               \\ \midrule
				
				\multirow{4}{*}{MixBlockA} 
				&256$\times$56$\times$56 &216 & 3195      & 27.1  &7.9        & 117           \\ 
				&512$\times$28$\times$28  &860 & 989       & 4.80   &182       & 206            \\ 
				& 1024$\times$14$\times$14 & 3400 & 712    & 1.56  &2200    & 456               \\ 
				& 2048$\times$7$\times$7   & 13710& 676    & 1.08  &\textbf{12694}  &\textbf{625}             \\  \midrule
				
				\multirow{4}{*}{MixBlockB/C} 
				&256$\times$56$\times$56 &216 & 3195      & 26.1      &8.1   & 122           \\ 
				&512$\times$28$\times$28  &860 & 989       & 4.50    & 191  & 219              \\ 
				& 1024$\times$14$\times$14   &3400 & 712    & 1.51    &2288   & 474             \\ 
				& 2048$\times$7$\times$7   &13710  & 676      & 1.05  &\textbf{13057}&\textbf{644}              \\ 
				
				\bottomrule
		\end{tabular}}
		\label{tab:Complexities}
	\end{table}
	
	\textbf{Guideline 1: Transformer block at the late stages maximizes the trade-off between efficiency and performance.}
	It is believed that Transformer block is more powerful but less efficient than the convolution block because the Transformer aims to build a global connection between features. In contrast, convolution only captures the local information \cite{Swin, Twins}. Here, we provide a detailed analysis equipped with TensorRT. As verified from Figure~\ref{result}, stacking Transformer blocks(PVT-Medium) gives $81.2\%$ $\text{Top-1}$ accuracy on ImageNet1K classification task while the corresponding accuracy of BottleNeck blocks(ResNet50) is only $79.0\%$.  However, as shown in Table \ref{tab:Complexities}, when it comes to efficiency, BottleNeck block always outperforms Transformer block in terms of TeraFLOPS under various input resolutions.
	
	Furthermore, the efficiency gap between Transformer and BottleNeck decreases when the input resolution decreases. Specifically, TeraFLOPS value of Transformer is only 81, less than quarter of the BottleNeck's TeraFLOPS value 378, when the input resolution is $56 \times 56$. However, when the resolution decreases to $7 \times 7$, Transformer's TeraFLOPS value is almost equal to BottleNeck, which is 599 and 670, respectively. This observation motivates us to put Transformer blocks at the late stages to balance performance and efficiency when designing efficient networks on TensorRT. Although some researchers(e.g., CoATNet\cite{Coatnet}) follow similar rules and propose networks mixed with Transformer and CNN blocks, to our best knowledge, we are the first to provide quantitative analysis on TensorRT.
	
	Following guideline 1, a native solution is to replace BottleNeck blocks with Transformer blocks at the late stages, dubbed as \text{MixNetV} (MixNet Valilla) in Table \ref{tab:Complexities}. We can see that MixNetV is faster than ViT and more accurate than ResNet.

	\textbf{Guideline 2: shallow-then-deep stage pattern improves the performance.} It is a widely held view that the more parameters, the higher the model capacity. We aim to develop a stage pattern that owns more parameters without sacrificing efficiency.  We first define the parameter density of an operation or block to quantify its capacity on hardware, similar to TeraFLOPs in Equ. \ref{eq:TeraFLOPS}, as follows: 
	\begin{equation}
	\text{TeraParams} = \frac{\text{Params}}{\text{Latency}}
	\label{eq:TeraParams}
	\end{equation}
	
	We study the property of TeraParams in various SOTA networks, e.g., RestNet and ViT, and present our findings in Table \ref{tab:Complexities}. The TeraParams of BottleNeck are getting more prominent in the late stages, which indicates that stacking more BottleNeck blocks at the late stage will result in larger model capacity, compared to the early stage. A similar trend can also be found in Transformer block in ViT.
	This observation inspires us to make the early stages shallower and the late stages deeper. We assume this shallow-then-deep stage pattern will bring more parameters, improve the model capacity and result in better performance without degrading the efficiency. To verify this, we make corresponding modifications on ResNet50 and provide the empirical results in Table \ref{stage_depth_comparsion}. We reduce the stage depth of the first and second stages from 3, 4 to 2, 3, respectively, while increasing the stage depth of the last stage from 3 to 5. From Table \ref{stage_depth_comparsion}, we can see that Refined-ResNet50 outperforms ResNet50 by $0.3\%$ Top-1 accuracy on ImageNet1K while being slightly faster on TensorRT. Furthermore, we apply this shallow-then-deep stage pattern to PvT-Small and get Refined-PvT-Small, which is better and faster. Finally, a similar conclusion can be made on MixNetV, which comes as no surprise.
	
	\begin{table}[h]
		\centering\caption{Stage depth comparsion.  Latency is measured on the TensorRT framework with a T4 GPU (batch size=8).}
		\label{stage_depth_comparsion}
		\resizebox{0.9\textwidth}{!}{
			\begin{tabular}{c|c|c|c|c|c}
				\toprule
				Model        	& Stage depth         & FLOPs(G)  & Param(M) & Latency(ms) & Top-1 Acc(\%)  \\ \midrule
				ResNet50\cite{ResNet}                &3-4-6-3       & 4.1    & 25.6  & 4.6        & 79.0  \\ 
				Refined-ResNet50                   &2-3-6-5           &4.1        & 34.1   &4.4        &79.3   \\ \midrule
				
				PVT-Small \cite{PVT_v1}                  &3-4-6-3     & 3.8    &24.5   & 20.7      &79.8 \\  
				Refined-PVT-Small                           &2-3-6-5    &3.6      &29.8   & 17.9      &  80.2  \\   \midrule
				
				MixNetV                &	3-5-6-3  & 4.3      &53.5      & 6.9       & 80.5      \\ %
				Refined-MixNetV  &2-3-6-4      &4.8        & 69.6     &    6.8     & 80.8        \\ 
				\bottomrule
		\end{tabular}}

	\end{table}
	
	\begin{figure}[h]
		\centering
		\includegraphics[scale=0.33]{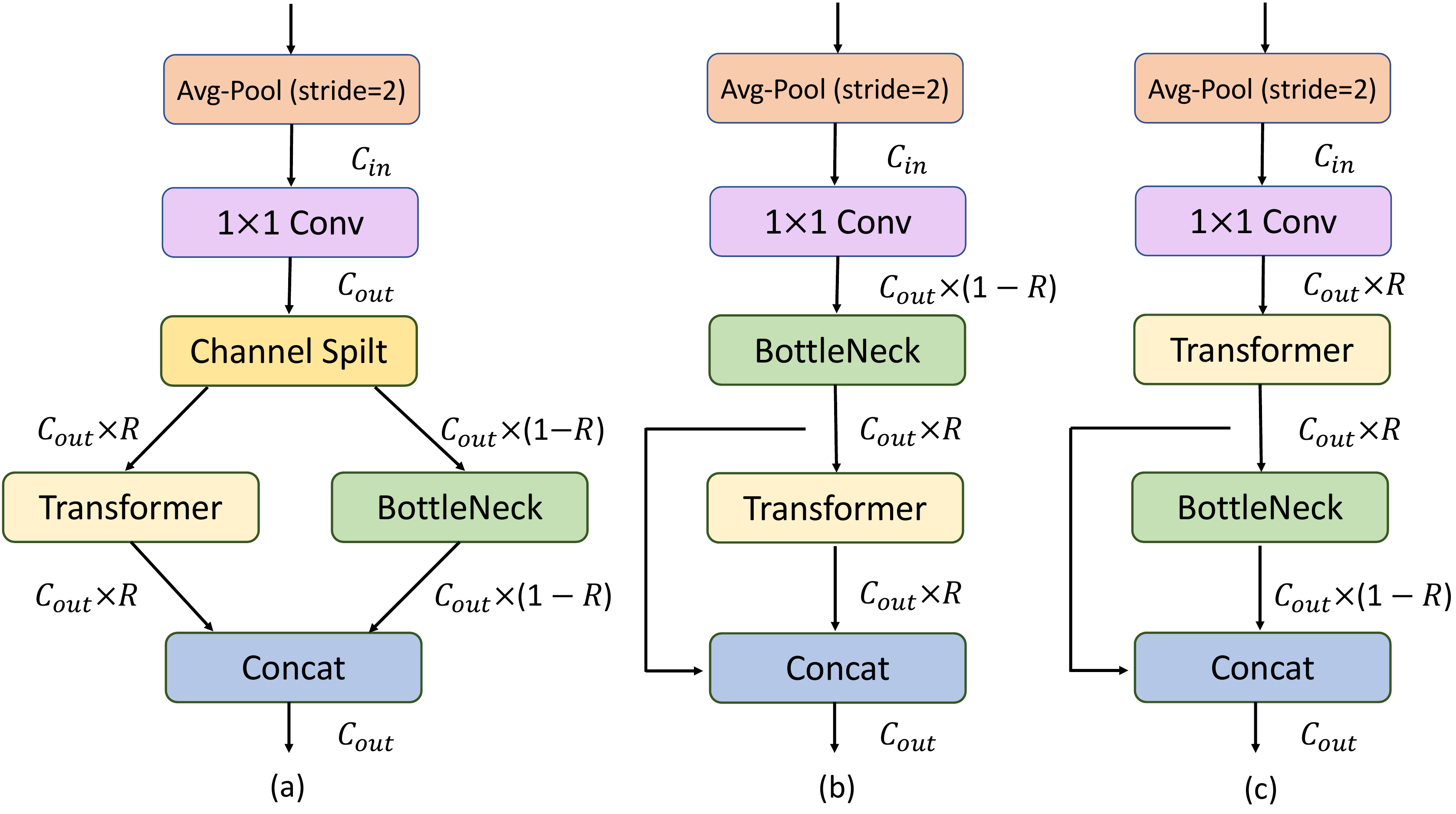}
		\caption{The proposed TRT-ViT block. (a): MixBlockA. (b): MixBlockB (c): MixBlockC. We use avgpool to downsamle while $\text{stride}=2$.} 
		\label{figure:block}
	\end{figure}
	
	\textbf{Guideline 3: a block mixed of Transformer and BottleNeck is more efficient than Transformer.}  Although we have achieved significant improvement following Guidelines 1 and 2, Refined-MixNetV can not dominate ResNet in efficiency. Our goal is to build a network with ResNet-like efficiency and ViT-like performance. To achieve this, we seek to design a new block mixed of Transformer and BottleNeck by leveraging the benefits of both. 
	
	We provide two mixed blocks in Figure \ref{figure:block}(a) and (b) where a $1\times1 \text{Conv}$ layer is first utilized to project the number of input channels, and a \text{Concat} layer is used to merge the global and local feature at the end. A shrinking ratio $R$ is introduced to control the proportion of Transformer block in the mixed block, defined as the output channel of Transformer block divided by $C_{out}$, the output channel of the mixed block. As depicted in Figure\ref{figure:block}(a), in MixBlockA, two branches are created, one for Transformer and the other for BottleNeck. In addition, a channel split layer is used to improve further the efficiency, where the output channel for Transformer block is $C_{out} \times R$ while the corresponding one is $C_{out} \times (1-R)$. In MixBlockB, Transformer block and BottleNeck block are stacked sequentially and the output channel of Transformer block and BottleNeck block are equal, i.e., $R=0.5$ here. As we can see from Table~\ref{tab:Complexities}, TeraFLOPS of both mixed blocks are significantly increased compared to Transformer block, which indicates the mixed blocks are more efficient than Transformer block. Moreover, their corresponding TeraParams are also better, which shows they have the potential to achieve better performance. 
	
	Replacing Transformer block in MixNetV with MixBlockA and MixBlockB, we build MixNetA and MixNetB. As depicted in Table~\ref{block_type_comparsion}, where C, T and M denote \underline{C}onvolution, \underline{T}ransformer and \underline{M}ixBlock respectively, we can see that both MixNetA and MixNetB outperform ResNet50. However, neither of them can beat MixNetV in terms of accuracy. Moreover, the performance gap between PVT-Medium and both mixed nets can not be neglected.
	
	

	\begin{table}[h]
		\centering\caption{Blcok type comparsion. Latency is measured on the TensorRT framework with a T4 GPU (batch size=8).}
		\label{block_type_comparsion}
		\resizebox{0.8\textwidth}{!}{
			\begin{tabular}{c|c|c|c|c}
				\toprule
				Model &	Block-type         & FLOPs(G) & Latency(ms) & Top-1 Acc(\%)  \\ \midrule
				ResNet50	&$\text{C-C-C-C}$       & 4.1       & 4.6        & 79.0  \\  
				PVT-Medium	&$\text{T-T-T-T}$      & 6.7       &30.0       & 81.2  \\  
				MixNetV	&$\text{C-C-C-T}$      & 4.8       & 6.8      & 80.8  \\  \midrule
				MixNetA	&$\text{C-C-C-M}$      & 2.7       & 4.3       & 80.7  \\ 
				MixNetB	&$\text{C-C-C-M}$      & 2.7       & 4.1       & 80.3 \\ 
				MixNetC	&$\text{C-C-C-M}$      & 2.7       & 4.1       & 81.2    \\ 
				\bottomrule
		\end{tabular}}
	\end{table}
	
	\textbf{Guideline 4: global-then-local block pattern helps to make up the performance issue.} As described in many papers \cite{Swin, PVT_v1, Twins}, Transformer whose receptive field is usually huge extracts global information from input features and enables information exchange within each entry. In contrast, Convolution, whose small receptive field only mines local information. It makes more sense to acquire global information first and refine it locally instead of extracting local information first and refining it globally. Moreover, the shrinking ratio $R$ in MixBlockB has to be 0.5, losing the flexibility to make a further adjustment. The global-then-local block pattern solves this issue, where $R$ can be any value between 0 and 1.
	
	Following this rule, we switch the order of Transformer operation and Convolution operation in MixBlockB and get MixBlockC, as depicted in Figure \ref{figure:block}(c). The TeraFLOPS score of MixBlockC is equivalent to MixBlockB. Then, substituting MixBlockB with MixBlockC, we get MixNetC. In coincidence with our intuition,  as shown in Table~\ref{block_type_comparsion}, MixNetC performs better than MixNetB without any sacrifice of efficiency. Moreover, MixNetC surpasses MixNetV in terms of both accuracy and TRT latency. Besides, the gap between MixNetC and PVT-Medium is bridged and MixNetC achieves ResNet-like efficiency and Transformer-like performance.
	
	We use MixBlockC as the basic building block in the rest of the paper.

	\subsection{TRT-ViT: An Efficient Network on TensorRT}
	We build our TRT-ViT architecture following the basic configuration of ResNet \cite{ResNet}, where feature pyramid structure is adopted and the resolution of feature maps decreases with network depth while the number of channels increases. Next, we apply our proposed practical guidelines to construct TRT-ViT. The whole architecture is divided into five stages, and we only use MixBlockC in the late stages, while convolution layers are used in the early stages. Besides, we utilize a shallow-then-deep stage pattern, where early stages are shallower and late stages are deeper, compared to the stage pattern in ResNet. 
	
	To provide a fair comparison with other SOTA networks, we propose four variants, TRT-ViT-A/B/C/D, whose configurations are listed in Table\ref{tab:structure}. $3\times3$ denotes convolution with kernel size =3 and C represents the output channel of a stage. $R, S, K$ are inter parameters in the proposed MixBlockC, where $S$ stands for spatial reduction ratio in Transformer Block\cite{PVT_v1}, $K$ indicates kernel size of BottleNeck in MixBlockC, and $R$ means the shrinking ratio. For simplicity, we set $R=0.5$. The stage depth of TRT-ViT-A/B/C/D is 2-4-5-4, 3-4-7-4, 3-4-9-6, 4-5-9-5, respectively. For all variants, we use MixBlockC in the final stage, i.e., stage5. For the larger model, TRT-ViT-C/D, we additionally use two MixBlcokC at stage4. Additionally, the expansion ratio of each MLP layer is set as 3, the head dim in Transformer is set as 32. For normlization and activation function, BottleNeck block use BatchNorm\cite{Batch_Norm} and ReLU, Transformer block use LayerNorm\cite{Layer_Norm} and GeLU\cite{GeLU}.

	\begin{table}[h]
		\centering
		\caption{Detailed configurations of our TRT-ViT variants in different stages}
		\label{tab:structure}
		\resizebox{0.9\textwidth}{!}{
			\begin{tabular}{@{}c|c|c|c|c|c@{}}
				
				\toprule
				Stage         &  Output size       & TRT-ViT-A             & TRT-ViT-B        & TRT-ViT-C   & TRT-ViT-D                 \\ \midrule
				
				\multirow{2}{*}{\makecell{Stem}} & \multirow{2}{*}{\makecell{$\displaystyle{\frac{H}{2}}\times \displaystyle{\frac{W}{2}}$}} &
				$C=32$ & $C=32$ & $C=32$ & $C=32$ \\
				\cmidrule{3-6} &    &
				$3\times3$ &
				$3\times3$ &
				$3\times3$ &
				$3\times3$ \\ \midrule
				
				\multirow{4}{*}{\makecell{Stage1}} & \multirow{3}{*}{\makecell{$\displaystyle{\frac{H}{2}}\times \displaystyle{\frac{W}{2}}$}} &
				$C=64$ & $C=64$ &$C=64$& $C=64$ \\
				\cmidrule{3-6} &    &
				$\begin{bmatrix}\ 3\times3 \\3\times3 \\   \end{bmatrix} \times1$ &
				$\begin{bmatrix}\ 3\times3\\ 3\times3 \\  \end{bmatrix} \times1$ &
				$\begin{bmatrix}\ 3\times3\\ 3\times3 \\  \end{bmatrix} \times1$ &
				$\begin{bmatrix}\ 3\times3\\ 3\times3 \\  \end{bmatrix} \times1$ \\ \midrule
				
				\multirow{5}{*}{\makecell{Stage2}} & \multirow{5}{*}{\makecell{$\displaystyle{\frac{H}{4}}\times \displaystyle{\frac{W}{4}}$}} &
				$C=160$ & $C=192$& $C=192$ & $C=256$ \\
				\cmidrule{3-6} &    &
				$\begin{bmatrix}\ 1\times1\\ 3\times3 \\ 1\times1 \\ \end{bmatrix} \times2$ &
				$\begin{bmatrix}\ 1\times1\\ 3\times3 \\ 1\times1 \\ \end{bmatrix} \times3$ &
				$\begin{bmatrix}\ 1\times1\\ 3\times3 \\ 1\times1 \\ \end{bmatrix} \times3$ &
				$\begin{bmatrix}\ 1\times1\\ 3\times3 \\ 1\times1 \\ \end{bmatrix} \times4$ \\ \midrule
				
				\multirow{5}{*}{\makecell{Stage3}} & \multirow{5}{*}{\makecell{$\displaystyle{\frac{H}{8}}\times \displaystyle{\frac{W}{8}}$}} &
				$C=320$ & $C=384$ & $C=384$ & $C=512$ \\
				\cmidrule{3-6} &    &
				$\begin{bmatrix}\ 1\times1\\ 3\times3 \\ 1\times1 \\ \end{bmatrix} \times4$ &
				$\begin{bmatrix}\ 1\times1\\ 3\times3 \\ 1\times1 \\ \end{bmatrix} \times4$ &
				$\begin{bmatrix}\ 1\times1\\ 3\times3 \\ 1\times1 \\ \end{bmatrix} \times4$ &
				$\begin{bmatrix}\ 1\times1\\ 3\times3 \\ 1\times1 \\ \end{bmatrix} \times5$ \\ \midrule
				
				\multirow{5}{*}{\makecell{Stage4}} & \multirow{5}{*}{\makecell{$\displaystyle{\frac{H}{16}}\times \displaystyle{\frac{W}{16}}$}} &
				$C=640$ &$C=768$ &$C=768$ & $C=1024$ \\
				\cmidrule{3-6} &    &
				$\begin{bmatrix}\ 1\times1\\ 3\times3 \\ 1\times1 \\ \end{bmatrix} \times5$ &
				$\begin{bmatrix}\ 1\times1\\ 3\times3 \\ 1\times1 \\ \end{bmatrix} \times7$ &
				$\begin{bmatrix}\ 1\times1\\ 3\times3 \\ 1\times1 \\ \end{bmatrix} \times7 + \begin{bmatrix}\ R=0.5\\\ S=2 \\ K=7 \\ \end{bmatrix} \times2$ &
				$\begin{bmatrix}\ 1\times1\\ 3\times3 \\ 1\times1 \\ \end{bmatrix} \times7 + \begin{bmatrix}\ R=0.5\\\ S=2 \\ K=7 \\ \end{bmatrix} \times2$ \\ \midrule
				
				\multirow{5}{*}{\makecell{Stage5}} & \multirow{5}{*}{\makecell{$\displaystyle{\frac{H}{32}}\times \displaystyle{\frac{W}{32}}$}} &
				$C=1280$ &$C=1536$ & $C=1536$ & $C=2048$ \\
				\cmidrule{3-6} &    &
				$\begin{bmatrix}\ R=0.5\\ S=1 \\ K=7 \\ \end{bmatrix} \times4$ &
				$\begin{bmatrix}\ R=0.5\\ S=1 \\ K=7 \\ \end{bmatrix} \times4$ &
				$\begin{bmatrix}\ R=0.5\\ S=1 \\ K=7 \\ \end{bmatrix} \times6$ &
				$\begin{bmatrix}\ R=0.5\\\ S=1 \\ K=7 \\ \end{bmatrix} \times5$ \\ 
				
				\bottomrule
		\end{tabular}}
	\end{table}


	\section{Experimental Results}

	\subsection{ImageNet-1K Classification}
	\noindent\textbf{Implementation Details.}  
	We first present the image classification on the ImageNet-1K \cite{ImageNet-1K}, which contains about 1.28M training images and 50K validation images from 1K categories.
	For a fair comparison, we follow the training settings of the recent vision Transformer \cite{Twins, Swin}.
	Concretely, all TRT-ViT variants are trained for 300 epochs on 8 V100 GPUs with a total batch size of 1024. The resolution of the input image is resized to 224 $\times$ 224.
	We adopt the AdamW\cite{AdamW} as the optimizer with weight decay 0.05.
	The learning rate is gradually decayed based on the cosine strategy with the initialization of 0.001 and uses a linear warm-up strategy with 30 epochs.
	Besides, we have also employed the stochastic depth\cite{Stochasticdepth} with the maximum drop-path rate of 0.1 for our TRT-ViT models. For latency measurement, we use TensorRT-8.0.3 on a T4 GPU with batch size=8.
\begin{table}[h]
	\centering
	\caption{Comparison of different state-of-the-art methods on ImageNet-1K classification. Latency is tested based on the TensorRT-8.0.3 with a T4 GPU (batch size=8)}
	\label{table:result_ImageNet}
	\centering
	{
		\scalebox{0.78}{
			\begin{tabular}{c|ccc|c}
				\toprule
				\multirow{2}{*}{Method}                     & Param     & FLOPs         & Latency         & Top-1 Acc  \\ 
				& (M)       & (G)           & (ms)            & (\%) \\ \midrule
				\multicolumn{5}{c}{ConvNet} \\   \midrule
				ResNet50\cite{ResNet}                     & 25.6      &  4.1            & 4.6                & 79.0           \\
				ResNet101\cite{ResNet}                     & 44.6     & 7.9            & 7.4                & 80.8          \\
				ResNet152\cite{ResNet}                    & 60.2      & 11.5            & 11.5                & 81.7          \\ 
				ConvNeXt-T\cite{ConvNext}                &29.0       & 4.5            & 18.1                & 82.1          \\ 
				ConvNeXt-S\cite{ConvNext}                &50.0       & 8.7            & 28.3                & 83.1          \\ 
				EfficientNet-B0 \cite{EfficientNet}          & 5.3      &  0.4         & 4.9                & 77.1          \\
				EfficientNet-B1 \cite{ResNet}                & 7.8     & 0.7            & 8.1           & 79.1         \\
				EfficientNet-B2 \cite{ResNet}                 & 9.2      & 1.0            & 9.0           & 80.1        \\
				EfficientNet-B3 \cite{ResNet}                    &12     & 1.8            & 18.2         & 81.6          \\ 
				RegNetY-8G\cite{RegNet}                     & 39.0      & 8.0           & 12.4              & 81.7          \\
				RegNetY-16G\cite{RegNet}                    & 84.0      & 16.0          & 18.4                & 82.9          \\ 
				\midrule
				
				\multicolumn{5}{c}{Transformer}                                                                            \\   \midrule

				T2T-ViT-14\cite{T2T}                        & 22.0      & 5.2          & -                & 81.5        \\
				CvT-13\cite{CvT}                            & 20.0      & 4.5          & -                & 81.6        \\
				Swin-T\cite{Swin}                           & 29.0      & 4.5           & -                & 81.3        \\
				LeViT-192\cite{LeViT}                       & 10.9      & 0.7             & 4.0            & 80.0        \\ 
				DeiT-Small/16$_{224}$\cite{Deit}        & 22.0      & 4.6          & 11.5               & 79.9        \\
				ViT-Small/32$_{384}$\cite{Deit}          &  22.9    & 3.5          & 12.6             & 80.0       \\
				PiT-S\cite{PiT}                           & 23.5     & 1.3           &13.8                & 80.9        \\
				Twins-SVT-S\cite{Twins}                     & 24.0      & 2.9             & 18.8             & 81.7        \\ 
				PVT-Small\cite{PVT_v1}                      & 24.5      & 3.8             & 20.7            & 79.8        \\
				Twins-PCPVT-S\cite{Twins}                   & 24.1      & 3.8             & 20.2             & 81.7        \\ 
				PVT-V2-B1\cite{PVT_v1}                      & 13.1      & 2.1             &21.1            & 78.7       \\
				CMT-XS\cite{Cmt}                         & 18.9      & 1.5             &  23.2          & 81.8        \\ 		
				CoaT-Lite-Small\cite{CoaT}                  & 20.0      & 4.0           & 29.8                & 81.9        \\
				TNT-S\cite{TNT}                             & 23.8      & 5.2          & 32.5              & 81.3        \\
				
				\textbf{TRT-ViT-A}                          &29.3         & 2.7      & \textbf{4.1}   &\textbf{81.2}  
				\\  
				\textbf{TRT-ViT-B}                           & 43.1    &  4.0          & \textbf{5.6}          & \textbf{82.1}  
				\\  \bottomrule

				CSWin-T\cite{CSWin}                         & 23.0      & 4.3             & 30.2            & 82.7        \\
				Twins-PCPVT-B\cite{Twins}                   & 43.8      & 6.7             & 29.1            & 82.7        \\ 
				PVT-Medium\cite{PVT_v1}                     & 44.2      & 6.7             & 30.0             & 81.2        \\
				PVT-v2-B2\cite{PVT_v2}                      & 25.4      & 4.0             & 37.8             & 82.0        \\
				XCiT-M24/16\cite{Xcit}                        & 84.4      & 16.2             & 41.2            & 82.6        \\
				SepViT-T\cite{SepViT}                       & 31.2      & 4.5             & 23.7            & 82.5        \\ 
				\textbf{TRT-ViT-C}                          &  67.3    &   5.9         & \textbf{9.2}             & \textbf{82.7}        \\  
				
				\midrule
				
				PVT-Large\cite{PVT_v1}                      & 61.4      & 9.8             & 43.2             & 81.7 \\
				CSWin-S\cite{CSWin}                         & 35.0      & 6.9             & 52.0           & 83.6 \\
				CMT-S\cite{Cmt}                         & 25.1      & 4.0             &  37.4          & 83.5        \\ 	
				PVT-v2-B3\cite{PVT_v2}                      & 45.2      & 6.9             & 54.1            & 83.2 \\
				SepViT-S\cite{SepViT}                       & 46.6      & 7.5             & 34.5            & 83.5 \\  
				Twins-PCPVT-L\cite{Twins}                   & 60.9      & 9.8             & 39.8             & 83.1 \\ 
				Twins-SVT-B\cite{Twins}                     & 56.0      & 8.6             & 38.9            & 83.2 \\
				\textbf{TRT-ViT-D}                           & 103.0           &9.7                & \textbf{15.1}  & \textbf{83.4} \\  
				\bottomrule
		\end{tabular}}
	}
\vspace{-0.2cm}
\end{table}

	\noindent\textbf{Comparison with State-of-the-art Models.}  As shown in Table \ref{table:result_ImageNet}, compared to the latest state-of-the-art methods, we achieve the best trade-off between accuracy and latency.
	Specifically, TRT-ViT-A achieves 81.2\% top-1 accuracy, 2.2\% higher than ResNet50  with about 10\% lower latency. TRT-ViT-B achieves similar accuracy compared with recent SOTA models ConvNeXt-T while 2.3$\times$ faster.
	Moreover, TRT-ViT-C outperforms the PVT-Medium by 1.5\% with the 2.2$\times$ faster speed.  Furthermore, compared to the recent models SepViT-T, TRT-ViT-C also achieves slightly better results while the inference speed is increased by 1.5$\times$. 
	When it comes to larger variant, TRT-ViT-D costs about 70\% fewer inference time than CSWin-S with similar performance. These results demonstrate that the proposed TensorRT-oriented Vision  Transformer design is an effective and promising paradigm.
	
	\subsection{ADE20K Semantic Segmentation}
	
	\noindent\textbf{Implementation Details.}
	To further verify the robustness of our TRT-ViT, we conduct the semantic segmentation experiment on ADE20K \cite{ADE20K}, which contains about 20K training images and 2K validation images from 150 categories.
	For fair comparison, we follow the training strategy of the previous vision Transformers \cite{PVT_v1,Swin,Twins,SepViT} on the Semantic FPN \cite{Semantic_FPN} and UperNet \cite{UperNet} frameworks. All of our models are pre-trained on the ImageNet-1k and then finetuned on ADE20K with the input size of 512$\times$512. 
	Based on the Semantic FPN framework, we adopt the AdamW optimizer with both the weight decay and learning rate being 0.0001. Then we train the whole network for 80K iterations with a total batch size of 16 and the stochastic depth of 0.1.
	For the training and testing on the UperNet framework, we train the models for 160K iterations. AdamW optimizer is used as well but with the learning rate $6\times10^{-5}$, total batch size 16, stochastic depth of 0.1, and weight decay 0.01. Then we test the mIoU based on both single-scale and multi-scale (MS), where the scale goes from 0.5 to 1.75 with an interval of 0.25.
	
	\noindent\textbf{Comparison with State-of-the-art Models.}
	In Table \ref{tab:ADE20K}, we make a comparison with the recent Transformer and CNN backbones.
	Based on the Semantic FPN framework,  TRT-ViT-B surpasses Twins-SVT-S by 2.2\% with about 83\% lower latency.
	Meanwhile, TRT-ViT shows great advantage over CNNs (e.g., ResNet\cite{ResNet}, ResNeXt\cite{ResNeXt}). Specifically, TRT-ViT-A achieves 44.2\% mIoU, which surpasses ResNet101 by 5.4\% but still faster(from 114ms to 82ms). For the UperNet framework, TRT-ViT-C achieves 2.2\% higher MS mIoU than recent SOTA CNN model ConvNeXt-T with 17\% fewer inference time. TRT-ViT-D achieves similar accuracy with Twins-SVT-L while much faster (from 1632ms to 301ms). Extensive experiments reveal that our TRT-ViT achieves excellent potential on segmentation tasks.
	\begin{table}[h]
		\centering
		\caption{Comparison of different backbones on ADE20K semantic segmentation task. FLOPs are measured with the input size of 512$\times$2048. Latency is backbone tested with the input size of 512$\times$2048 based on the TensorRT framework in T4 GPU (batch size=8). }
		\label{tab:ADE20K}
		\resizebox{0.95\textwidth}{!}{
			\begin{tabular}{c|c|ccc|ccc}
				\toprule
				\multirow{2.5}{*}{Backbone}  &\multirow{2.5}{*}{Latency(ms)}     & \multicolumn{3}{c|}{Semantic FPN 80k}   & \multicolumn{3}{c}{UperNet 160k}          \\ \cmidrule(l){3-8}
				&  & Param(M)   & FLOPs(G)   & mIoU(\%)      & Param(M)  & FLOPs(G)    & mIoU/MS mIoU(\%)  \\   \midrule
				ResNet50\cite{ResNet}          &72  & 28.5       & 183        & 36.7          & -          & -          & -/-            \\
				ResNet101\cite{ResNet}        &114   & 47.5       & 260        & 38.8          & 86.0       & 1092       & -/44.9         \\
				PVT-Small\cite{PVT_v1}         &445  & 28.2       & 161        & 39.8          & -          & -          & -/-            \\
				Twins-SVT-S\cite{Twins}        &636  & 28.3       & 144        & 43.2          & 54.4       & 901        & 46.2/47.1      \\
				\textbf{TRT-ViT-A}            & 82   &  32.1            &146         &44.2           &  65.5           &  909           & 45.3/46.2      \\
				\textbf{TRT-ViT-B}            &  104  &    46.4         &  176          &	\textbf{45.4}           &   80.9     &  941            & \textbf{46.5}/\textbf{47.5}      \\
				\midrule

				ResNeXt101-64$\times$4d\cite{ResNeXt} &231 & 86.4    & -          & 40.2          & -          & -          & -/-            \\
				
				PVT-Medium\cite{PVT_v1}       &672   & 48.0       & 219        & 41.6          & -          & -          & -/-            \\
				ConvNeXt-T\cite{Twins}       &228   & -     & -        & -          & 60.0     & 939       & -/46.7      \\
				Twins-PCPVT-B\cite{Twins}     &625   & 48.1       & 220        & 44.9          & 74.3       & 977        & 47.1/48.4      \\
				PVT-Large\cite{PVT_v1}        & 916  & 65.1       & 283        & 42.1          & -          & -          & -/-            \\
				Twins-SVT-L\cite{Twins}       & 1632  & 103.7      & 404        & 46.7          & 133.0      & 1164       & 48.8/49.7      \\
				
				\textbf{TRT-ViT-C}          &  189    &70.6               &   213       &46.2           &  105.0    & 978          & 47.6/48.9      \\
				\textbf{TRT-ViT-D}            &  301  &   105.9          &   296         & \textbf{46.7}           & 143.7    &  1065      &  \textbf{48.8}/ \textbf{49.8}      \\
				
				\bottomrule
			\end{tabular}
		}

	\end{table}
	
	\subsection{COCO Object Detection and Instance Segmentation}

	\noindent\textbf{Implementation Details.}
	Next, we evaluate TRT-ViT on the objection detection and instance segmentation task \cite{COCO} based on the Mask R-CNN \cite{Mask_RCNN} frameworks with COCO2017 \cite{COCO}.
	Specifically, all of our models are pre-trained on ImageNet-1K and then finetuned following the settings of the previous works \cite{Twins,SepViT}.
	Based on the 12 epochs (1$\times$) experiment, we adopt the AdamW optimizer with the weight decay of 0.0001. There are 500 iterations for a warm-up during the training, and the learning rate will decline by 10$\times$ at epochs 8 and 11.
	Based on the 36 epochs (3$\times$) experiment with multi-scale (MS) training, models are trained with the resized images such that the shorter side ranges from 480 to 800 and the longer side is at most 1333. 
	Furthermore, the settings are the same as 1$\times$ except that the weight decay becomes 0.05, and the decay epochs are 27 and 33.

	\begin{table}[h]
		\centering
		\caption{Comparison of different backbones on Mask R-CNN-based framework. FLOPs are measured with the input size of $800 \times 1280$. Latency is the backbone tested with the input size of 800$\times$1280 based on the TensorRT framework in T4 GPU (batch size=8). The superscript $b$ and $m$ denote the box detection and mask instance segmentation. }
		\label{tab:MaskRCNN_COCO}
		\resizebox{0.99\textwidth}{!}{
			\begin{tabular}{c|c|c|c|cccccc|cccccc}
				\toprule
				\multirow{2}{*}{Backbone}  & Latency     & Param     & FLOPs     & \multicolumn{6}{c|}{Mask R-CNN 1$\times$}        & \multicolumn{6}{c}{Mask R-CNN 3$\times$ + MS}       \\ \cline{5-16} 
				& (ms)        & (M)       & (G)       & AP$^b$        & AP$_{50}^b$   & AP$_{75}^b$   & AP$^m$        & AP$_{50}^m$   & AP$_{75}^m$   & AP$^b$        & AP$_{50}^b$   & AP$_{75}^b$   & AP$^m$        & AP$_{50}^m$   & AP$_{75}^m$   \\ \midrule
				ResNet50\cite{ResNet}      &  70        & 44.2      & 260       & 38.0          & 58.6          & 41.4          & 34.4          & 55.1          & 36.7          & 41.0          & 61.7          & 44.9          & 37.1          & 58.4          & 40.1          \\
				ResNet101\cite{ResNet}   & 113      & 63.2      & 336       & 40.4          & 61.1          & 44.2          & 36.4          & 57.7          & 38.8          & 42.8          & 63.2          & 47.1          & 38.5          & 60.1          & 41.3          \\
				PVT-Small\cite{PVT_v1}     & 443     & 44.1      & 245       & 40.4          & 62.9          & 43.8          & 37.8          & 60.1          & 40.3          & 43.0          & 65.3          & 46.9          & 39.9          & 62.5          & 42.8          \\
				Twins-SVT-S\cite{Twins}   &628        & 44.0      & 228       & 43.4          & 66.0          & 47.3          & 40.3          & 63.2          & 43.4          & 46.8          & 69.2          & 51.2          & 42.6          & 66.3          & 45.8          \\
				
				\textbf{TRT-ViT-A}       &  86  & 48.4     & 229        & \textbf{42.4}          & \textbf{65.0}         & \textbf{46.2}          &\textbf{39.4}          & \textbf{62.2}          & \textbf{42.3}          & \textbf{45.4}          & \textbf{67.5}          & \textbf{49.7}          & \textbf{41.3}          & \textbf{64.6}          & \textbf{44.5}           \\
				\textbf{TRT-ViT-B}       & 110    & 62.1    & 258        & \textbf{43.6}          & \textbf{66.1}         & \textbf{47.5}          &\textbf{40.4}          & \textbf{63.7}          & \textbf{43.5}          & \textbf{46.9}          & \textbf{69.2}          & \textbf{51.3}          & \textbf{42.8}          & \textbf{66.5}          & \textbf{45.9}           \\
				\midrule
				
				ResNeXt101-32$\times$4d\cite{ResNeXt} & 131    & 63.0      & 340       & 41.9          & 62.5          & 45.9          & 37.5          & 59.4          & 40.2          & 44.0          & 64.4          & 48.0          & 39.2          & 61.4          & 41.9          \\
				PVT-Medium\cite{PVT_v1}    & 718          & 63.9      & 302       & 42.0          & 64.4          & 45.6          & 39.0          & 61.6          & 42.1          & 44.2          & 66.0          & 48.2          & 40.5          & 63.1          & 43.5          \\
				Twins-SVT-B\cite{Twins}   &  1255     & 76.3      & 340       & 45.2          & 67.6          & 49.3          & 41.5          & 64.5          & 44.8          & 48.0          & 69.5          & 52.7          & 43.0          & 66.8          & 46.6          \\
				ConvNeXt-T\cite{ConvNext}    &  235         & -      & 262         & -          & -          & -          & -          & -          & -          & 46.2            & 67.9        & 50.8         & 41.7            & 65.0          & 44.9             \\
				\textbf{TRT-ViT-C}   &  211        & 86.3      &294          &\textbf{44.7}          &\textbf{66.9}          & \textbf{48.8}         & \textbf{40.8}          & \textbf{63.9}          & \textbf{44.0}          & \textbf{47.3}         & \textbf{68.8}          &\textbf{51.9}          &\textbf{42.7}          &\textbf{65.9}         & \textbf{46.0}    \\ 
				\textbf{TRT-ViT-D}   & 332          &121.5      &375              &\textbf{45.3}          &\textbf{67.9}          & \textbf{49.6}         & \textbf{41.6}          & \textbf{64.7}          & \textbf{44.8}          & \textbf{48.1}         & \textbf{69.3}          &\textbf{52.7}          &\textbf{43.4}          &\textbf{66.7}         & \textbf{46.8}    \\ 
				\bottomrule
			\end{tabular}
		}

	\end{table}

	\noindent\textbf{Comparison with State-of-the-art Models.}
	Table \ref{tab:MaskRCNN_COCO} shows the evaluation result with the Mask R-CNN framework. Based on the 1$\times$ schedule, we can see that TRT-ViT-B achieves comparable performance with Twins SVT-S, while the inference speed is increased by $4.7\times$. Based on the 3$\times$ schedule, TRT-ViT-C achieves 47.3 AP, which is 1.1 higer than ConvNeXt-T with lower latency. TRT-ViT-D  achieves 48.1 AP, exhibits the similar AP as Twins-SVT-B with 2.7$\times$ faster speed.
	
	More importantly, as shown in Table\ref{tab:ADE20K} and Table\ref{tab:MaskRCNN_COCO},  Transformer backbone achieves even lower speed than CNN backbone when transferring to downstream tasks(large input size). For example, when transferring from classification to detection, the inference latency of Twins-SVT-B is increased by $32\times$ (from 38.9ms to 1255ms) while the inference latency of ResNet101 is only increased by $15\times$ (from 7.4ms to 113ms). There exists a speed gap between classification and downstream tasks in current Transformer. 
	Owing to the proposed approach, TRT-ViT still performs ResNet-like efficiency when transferring to downstream tasks, which implies TRT-ViT can bridge this gap.
	
	\subsection{Ablation Study}
	To understand our TRT-ViT better, we ablate each critical design by evaluating its performance on ImageNet-1K classification.  Latency is measured on the TensorRT framework with a T4 GPU (batch size=8).

	\noindent\textbf{Impact of different stage types.}
	We compare the accuracy of using Transformer Block at different stages, and the results are shown in Table \ref{tab:stage_type_comparsion}.  As stated in Table \ref{tab:stage_type_comparsion}, when we use Transformer block in an early stage, the latency increases significantly, but the accuracy only increases slightly, which verifies our first guideline.
	
	\begin{minipage}{\textwidth}
		\begin{minipage}[t]{0.5\textwidth}
			\makeatletter\def\@captype{table}
			\caption{ Stage type comparsion }
			\label{tab:stage_type_comparsion}
			\centering
			\resizebox{0.98\textwidth}{!}{
				\begin{tabular}{c|c|c|c}
					\toprule
					Stage-type & FLOPs(G) & Latency(ms) & Top-1 Acc(\%)  \\
					\midrule
					$\text{C-C-C-M}$ & 2.7 & \textbf{4.1}   & 81.2 \\ 
					$\text{C-C-M-M}$ & 3.7 & 9.1  & 81.6     \\ 
					$\text{C-M-M-M}$ & 4.5 & 15.2  & 81.8     \\ 
					\bottomrule
			\end{tabular}}
		\end{minipage}
		\begin{minipage}[t]{0.5\textwidth}
			\makeatletter\def\@captype{table}
			\caption{ Comparison of results of different ratios.}
			\label{tab:ratio_comparsion}
			\centering
			\resizebox{0.81\textwidth}{!}{
				\begin{tabular}{c|c|c|c}
					\toprule
					$R$       & FLOPs(G) & Latency(ms) & Top-1 Acc(\%)  \\ \midrule
					1.00            &   4.8           &   6.8                & 80.8    \\
					0.75          &   3.6            &   4.9                & 81.3    \\  
					0.50          &   2.7       &     4.1           &81.2     \\
					0.25          &   2.4            &  3.9           &80.6     \\

					\bottomrule
			\end{tabular}}
			
		\end{minipage}
	\end{minipage}

	\noindent\textbf{Impact of different ratio.}
	Furthermore, we compare the performance of different shrinking ratios in the MixBlockC. As stated in Table \ref{tab:ratio_comparsion}, when we decrease the shrinking ratio, the latency is reduced significantly with the cost of accuracy degradation, which is expected. Especially when the shrinking ratio $R$ is equal to 0.5,  the latency is compressed by 40\% and the performance is even slightly better than the baseline model$(R=1.0)$, indicating the superiority of the proposed TRT-ViT block.

	\section{Conclusion}
	\label{section:conclusion}
	In this paper, we directly treat the TensorRT latency as a computational efficiency metric to guide the model design, which provides a viable path to alleviate the gap between research and practical deployment.
	We conduct extensive experiments, summarize four practical guidelines for TensorRT-oriented network design, and propose a family of Transformers, abbreviated as TRT-ViT. 
	Experimental results demonstrate that TRT-ViT achieves a state-of-the-art latency/accuracy trade-off across diverse visual tasks, such as image classification, object detection and semantic segmentation.
	We hope that our work can provide inspiring insight and encourage more researchers to be involved in real-scene model design.

	{
		\small
		\bibliographystyle{plain}
		\bibliography{neurips_2022}
	}

	\appendix

	%
	%

\end{document}